# DEVELOPMENT OF AN AUTISM SCREENING CLASSIFICATION MODEL FOR TODDLERS

Afef Saihi and Hussam Alshraideh

Department of Industrial Engineering,
American University of Sharjah, Sharjah, UAE

## ABSTRACT

*Autism spectrum disorder ASD is a neurodevelopmental disorder associated with challenges in communication, social interaction, and repetitive behaviors. Getting a clear diagnosis for a child is necessary for starting early intervention and having access to therapy services. However, there are many barriers that hinder the screening of these kids for autism at an early stage which might delay further the access to therapeutic interventions. One promising direction for improving the efficiency and accuracy of ASD detection in toddlers is the use of machine learning techniques to build classifiers that serve the purpose. This paper contributes to this area and uses the data developed by Dr. Fadi Fayez Thabtah to train and test various machine learning classifiers for the early ASD screening. Based on various attributes, three models have been trained and compared which are Decision tree C4.5, Random Forest, and Neural Network. The three models provided very good accuracies based on testing data, however, it is the Neural Network that outperformed the other two models. This work contributes to the early screening of toddlers by helping identify those who have ASD traits and should pursue formal clinical diagnosis.*

## 1. INTRODUCTION

Autism spectrum disorder (ASD) is a neurodevelopmental condition that impacts the way a person perceives others and socializes with them. It affects three main developmental areas which are communication, social interaction, and repetitive patterns of behavior. As shown in Figure 1, Autism rates continue to rise dramatically, and according to Center for Disease Control (CDC) reports [1], prevalence rate increases with 1 in 54 children diagnosed with autism in 2020 compared to 1 in 59 in 2018, 1 in 110 in 2006, and 1 in 150 in 2000.

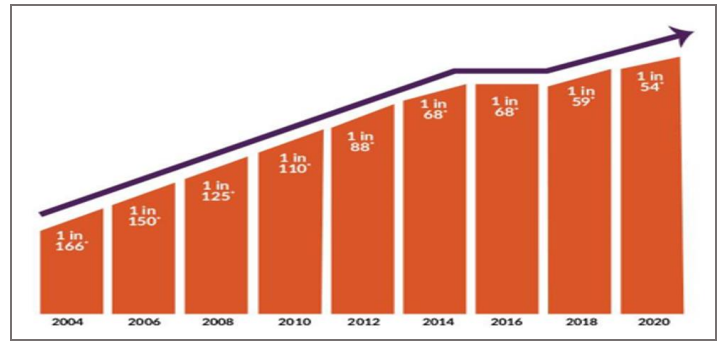

Figure 1. Autism prevalence estimates (Source: CDC reports [1])

Research has shown that there is no cure for ASD, however, early diagnosis and intensive intervention can make a big difference in the lives of many children and their families. Early diagnosis is very important for children on the spectrum because it allows to teach them the skills and behaviors that they lack at an early age when they still have good brain plasticity and therefore, the impact of intervention can be maximized and helps them reach their full potential. Interventions such as special education, behavior modification techniques, speech and occupational therapies help to bridge the gap that ASD kids have compared to their peers and speed up their development [2]. Therefore, early identification and diagnosis of children with ASD is very beneficial for the kids and their families and early screening is always recommended by experts because it allows to detect at an early stage the kids with more risk to be on the autism spectrum disorder, and thus, pushes their families to take the required measures to get the kids access the early intervention services.

Early screening of ASD has an important role in improving prognosis via early diagnosis and intervention [3]. For these kids with autism, both research and practice have shown that there is no magic pill for cure and early intervention through intensive education and behavior modification is the key due to its capability of changing the quality of life of these kids and their families and improving long-term outcomes [4]. However, the standardized tests for diagnosis such as Autism Diagnostic Observation Schedule (ADOS) and Autism Diagnostic Interview-Revised (ADI-R), among others are time consuming, very costly and can be run only by trained clinicians. Thus, there are many barriers of getting these kids screened for autism at an early stage which contribute to delaying the therapies and interventions that they require. Nowadays, due to the advancement made in artificial intelligence and machine learning, autism can be predicted at an incredibly early stage [5]. Having access to an accurate, automated, cost effective and fast instrument for ASD screening at an early age will be greatly beneficial for detecting autism traits in the kids and securing a much better future for them.

In this work, our aim is to contribute to the early screening of toddlers to help identify the kids who have ASD traits and should pursue formal clinical diagnosis. This is done through the development of a classification model that, based on some defined attributes, will identify the kids who have ASD symptoms and need to undergo further advanced assessments with professionals.

The remainder of this paper contains a literature review section that discusses the related works, a methodology section that describes the data and details the steps to be followed for the analysis, and a model building section, in which, the selected classifiers have been trained and their respective performances have been analyzed and compared.

## 2. LITERATURE REVIEW

Although concerns about children development milestones are mostly reported by parents of children with ASD within their first year, these children are rarely diagnosed before the age of 4 years [3, 6]. Early identification is key as, if followed by intensive early intervention, will enable these kids to acquire the necessary skills, improve the core behavioral symptoms, and there is a chance that they will grow out of the diagnosis or at least loose many of the autistic traits [7]. A significant number of studies investigated the potential for early intervention in helping kids with this neurodevelopmental condition. In this context, [8] conducted a systematic review that highlighted the large number of studies, which are over 83% of the published literature since 2010, reflecting the increased interest among researchers in this area.

Autism cannot be identified using conventional clinical methods such as blood tests. Instead, ASD screening is a crucial phase, and it is the process of determining the autistic symptoms of an individual. Many screening tools have emerged overtime, they include direct observations, questionnaires and interviews and they should be performed by specialists. However, these tools are lengthy, costly, and low-and-middle-income countries have shortage of mental health clinicians. This constitutes a major barrier for obtaining an early diagnosis and accessing intervention services [9]. Therefore, a viable screening instrument that is cost effective, available and less time consuming to identify the risk of ASD at a preliminary stage is highly needed.

There is a growing interest among researchers in the early screening and intervention for young children at risk of ASD [8] and in the use of machine learning and intelligent methods for autism screening and detection [9, 10]. In this context, [11] investigated fuzzy data mining models to detect autism features for both cases and control groups between 4 and 11 years. [5] developed an effective prediction model by merging Random Forest-CART and Random Forest-Id3 and complemented their work by developing a mobile application based on the proposed model. [12] developed and tested an Artificial Neural Network (ANN) for diagnosing ASD, and the test data evaluation showed that ANN model was able to diagnose ASD with 100% accuracy. [13] used machine learning to investigate the accuracy and reliability of the Q-CHAT method in classifying autistic kids. The authors used three different ML algorithms and found that the Support Vector Machine was the most effective. Similarly, [10] proposed a new machine learning method which is the "Rules-Machine Learning". This method detects the ASD traits in cases and offers rules that can be used by domain experts to understand the reasons behind the classification. The authors compared this method with other classifiers and found that it leads to higher predictive accuracy, sensitivity, and specificity than those of other models such as bagging, decision trees and rule induction. In line with these studies, this paper contributes to this area by proposing a toddler screening prediction model for ASD traits.

## 3. METHODOLOGY AND DESCRIPTION OF DATA

The dataset used in this study is obtained from Kaggle website, and it was developed by Dr Fadi Fayez Thabtah [14] to screen autism in toddlers. In this dataset, 10 behavioral features have been recorded in addition to some individual characteristics that are effective in detecting ASD cases. Table 1 summarizes the various attributes that the dataset includes.

Table 1. Dataset Description (source: [14])

| Variable | Type | Description |
|---|---|---|
| A1: Response_to_name | Binary (0, 1) | Does your child look at you when you call his/her name? |
| A2: Eye_contact | Binary (0, 1) | How easy is it for you to get eye contact with your child? |
| A3: Point_to_objects | Binary (0, 1) | Does your child point to indicate that s/he wants something? |
| A4: Sharing_interest | Binary (0, 1) | Does your child point to share interest with you? |
| A5: Pretend_play | Binary (0, 1) | Does your child pretend? |
| A6: Follow_looking | Binary (0, 1) | Does your child follow where you are looking? |
| A7: Confort_someone | Binary (0, 1) | does your child show signs of wanting to comfort someone upset? |
| A8: First_words | Binary (0, 1) | Description of child first words |
| A9: Simple_gesture | Binary (0, 1) | Does your child use simple gestures? |
| A10: Stare_at_nothing | Binary (0, 1) | Does your child stare at nothing with no apparent purpose? |
| Age | Number | Toddlers (months) |
| Score by Q-chat-10 | Number | 1-10 (Less than or equal 3 no ASD traits; > 3 ASD traits |
| Sex | Character | Male or Female |
| Ethnicity | String | List of common ethnicities in text format |
| Born with jaundice | Boolean (Y/N) | Whether the case was born with jaundice |
| Family member with ASD history | Boolean (Y/N) | Whether any immediate family member has a PDD |
| Who is completing the test | String | Parent, self, caregiver, medical staff, clinician, etc. |
| Class variable (ASD) | String | ASD traits or No ASD traits (Yes / No) |

The items A1 to A10 are answers to questions, the possible answers are Always, Usually, Sometimes, Rarely and Never. For the questions A1 to A9, 1 is assigned if the response was Sometimes or Rarely or Never, and 0 is assigned if the response was Always or Usually. However, for question 10, if the response was Always or Usually or Sometimes, 1 is assigned, otherwise 0 is assigned.

The proposed plan for analysis comprises the following steps:

1. Examining the data and its different attributes to see if any preprocessing steps are required before using the data with the prediction models.
2. Performing an exploratory data analysis to generate insights and hidden patterns from the data.
3. Providing some classification models such as Decision Trees, Ensemble methods, Neural networks that predict the ASD screening status based on the other provided attributes.
4. Comparing the various models based on the prediction accuracy and selecting the model that provides the best accuracy based on the validation data.
5. Evaluating the model using the testing dataset.

Throughout the analysis steps, R-Studio which is a development environment for R is used. Moreover, the dataset is divided into training dataset that is used for training and validation and testing data set that is used only at the very end to evaluate the model performance. The expected outcome of this study is to have an accurate prediction model that will help the autism community by contributing to screening the toddlers at an early stage.

## 4. ANALYSIS AND RESULTS

### 4.1. Exploratory Data Analysis

An initial investigation has been performed on the data to discover if any patterns exist and to generate insights. Table 2 summarizes the proportions of female vs male toddlers screened for ASD symptoms. Of the identified cases as having ASD traits, 26% are female and 74% are male which is in line with the research of [15] that analyzed fifty-four studies and found that, among children meeting criteria for ASD, the male-to-female ratio is close to three to one.

Table 2. Male vs female kids screened for ASD

| Sex | ASD traits | ASD traits | No ASD traits | Totals |
|---|---|---|---|---|
| **Female** | | 26.6% | 38.3% | 30.3% |
| **Male** | | 73.4% | 61.7% | 69.7% |
| **Totals** | | 100% | 100% | 100% |

The collected data contains cases from different ethnicities, and we are interested in investigating the distribution of the cases by ethnicity. Figure 2 provides a summary about this and shows that the majority of the toddlers having ASD traits are White European followed by Asian then Middle Eastern.

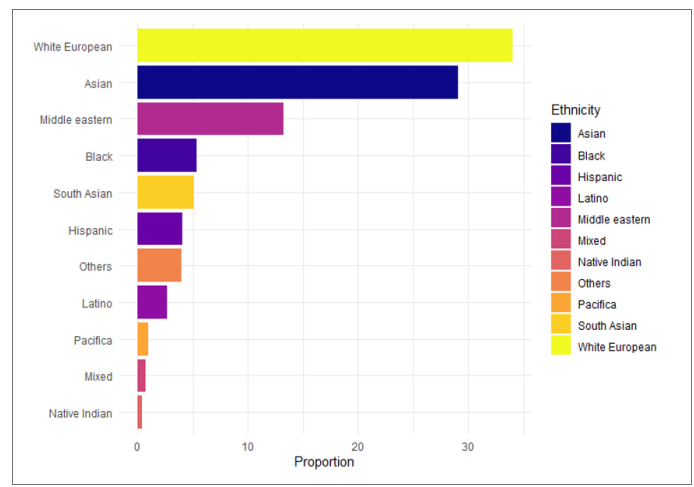

Figure 2. Proportion of ASD cases by ethnicity

### 4.2. Model Building

The ASD screening process is a binary classification problem as it aims to classify toddlers to either having or not having Autism traits [10]. We intend here to use various classification models, assess and compare their performances to see which model provides better performance measures and is therefore more suitable for this purpose. The evaluated models include Decision trees CART and C45, Ensemble methods Bagged cart and Random Forest, and Neural Networks. The outcome variable is the ASD.Traits (Yes/No) which indicates if the child is classified as having autism symptoms or no. This outcome is predicted based on the rest of the attributes presented in Table 1 except the score by Q-Chat-10 test, based on it, the data has been labelled. This variable is considered as an outcome variable and should not be included as predictor because this leads to model overfitting.

The classification models were built by performing a 5-fold cross-validation, using 70% of the dataset (739 observations) for training and the remaining 30% (315 observations) to test the accuracy of the various classifiers. Regarding the cross-validation, the training data set is partitioned into 5 subsets, and the algorithm randomly uses 4 data subsets to train the model and then tests it on the remaining subset.

### 4.2.1. Decision Tree Models

CART and C4.5 decision trees were used. Based on the validation dataset, CART tree gave an accuracy of 88% and C4.5 gave an accuracy of 92%. Table 3 presents the confusion matrices of both models. The plots of the two trees as well as the ROC curve of the C4.5 are in the Appendix. Given its better accuracy based on validation dataset, the C4.5 decision tree is a candidate for this category.

Table 3. Confusion matrices based on validation dataset

| **CART Tree** | | | | **C4.5 Tree** | | |
|---|---|---|---|---|---|---|
| | **Reference** | | | | **Reference** | |
| **Prediction** | No | Yes | | **Prediction** | No | Yes |
| No | 25 | 5.4 | | No | 27 | 3.9 |
| Yes | 6 | 63.7 | | Yes | 3.9 | 65.2 |
| Accuracy | 0.8861 | | | Accuracy | 0.9222 | |

The C4.5 model was applied to the test dataset (315 observations) to predict the outcome variable, and Table 4 summarizes its performance.

Table 4. C4.5 performance on testing data

| | **Reference** | | | | |
|---|---|---|---|---|---|
| **Prediction** | No | Yes | Sensitivity | 0.9908 |
| No | 25 | 5.4 | Specificity | 0.9794 |
| Yes | 6 | 63.7 | Pos Pred Value | 0.9908 |
| Accuracy | 0.9873 | | Neg Pred Value | 0.9794 |
| 95% CI | (0.9678, 0.9965) | | Prevalence | 0.6921 |
| No Information Rate | 0.6921 | | Detection Rate | 0.6857 |
| P-Value | <2e-16 | | Detection Prevalence | 0.6921 |
| Kappa | 0.9702 | | Balanced Accuracy | 0.9851 |
| | | | 'Positive' Class | Yes |

### 4.2.2. Ensemble Methods

From the Ensemble methods, Bagged CART and Random Forest models were used. Based on the validation dataset, Bagged CART gave an accuracy of 94% and Random Forest gave an accuracy of 95%. Table 5 presents the confusion matrices of both models. Therefore, the Random Forest is candidate for this category, and its ROC curve is presented in the Appendix.

Table 5. Confusion matrices based on validation dataset

| Bagged CART | | | | Random Forest | | |
|---|---|---|---|---|---|---|
| | Reference | | | | Reference | |
| **Prediction** | No | Yes | | **Prediction** | No | Yes |
| No | 28.1 | 2.6 | | No | 27.6 | 0.8 |
| Yes | 2.8 | 66.4 | | Yes | 3.4 | 68.2 |
| Accuracy | 0.9459 | | | Accuracy | 0.9581 | |

The Random Forest model was applied to the test dataset to predict the outcome variable, and Table 6 summarizes its performance on this data.

Table 6. Random Forest performance on testing data

| | Reference | | | | |
|---|---|---|---|---|---|
| **Prediction** | No | Yes | | Sensitivity | 1 |
| No | 85 | 0 | | Specificity | 0.8763 |
| Yes | 12 | 218 | | Pos Pred Value | 0.9478 |
| Accuracy | 0.9619 | | | Neg Pred Value | 1 |
| 95% CI | (0.9344, 0.9802) | | | Prevalence | 0.6921 |
| No Information Rate | 0.6921 | | | Detection Rate | 0.6921 |
| P-Value | <2.2e-16 | | | Detection Prevalence | 0.7302 |
| Kappa | 0.9074 | | | Balanced Accuracy | 0.9381 |
| | | | | 'Positive' Class | Yes |

### 4.2.3. Neural Networks

Artificial Neural Network (ANN) model was built using R Neuralnet package. The model has two hidden layers; the first one with 5 hidden nodes and the second one with 3 hidden nodes. The plot of the model is presented in the Appendix and Table 7 summarizes its performance on the testing data.

Table 7. ANN performance on testing data

| | Reference | | | | |
|---|---|---|---|---|---|
| **Prediction** | No | Yes | | Sensitivity | 0.9794 |
| No | 95 | 0 | | Specificity | 1 |
| Yes | 2 | 218 | | Pos Pred Value | 1 |
| Accuracy | 0.9937 | | | Neg Pred Value | 0.9909 |
| 95% CI | (0.9773, 0.9992) | | | Prevalence | 0.3079 |
| No Information Rate | 0.6921 | | | Detection Rate | 0.3016 |
| P-Value | <2e-16 | | | Detection Prevalence | 0.3016 |
| Kappa | 0.985 | | | Balanced Accuracy | 0.9897 |
| | | | | 'Positive' Class | No |

### 4.2.4. Models Comparison

The three candidate models are the C4.5, the Random Forest and the Artificial Neural Network. Table 8 summarizes the prediction accuracy, sensitivity, and specificity of these three models. All the candidate models have an extremely good performance and are expected to have an excellent predictive power in detecting autistic traits in toddlers.

Table 8. Models comparison

| Model | Accuracy | Sensitivity | Specificity |
|---|---|---|---|
| C4.5 Tree | 0.98 | 0.99 | 0.97 |
| Random Forest | 0.96 | 1 | 0.87 |
| Neural Network | 0.99 | 0.97 | 1 |

## 5. CONCLUSION

Due to the increasing rate of autism diagnosis cases and the multiple barriers that hinder getting a timely screening and therefore a diagnosis which allows the kids to get access to the early intervention services, there is an urgent need to think about developing and implementing a fast, accessible, and cost-effective autism screening tool. This was the aim of this work that showed the predictive power of the machine learning techniques in detecting autistic traits. The dataset used for training the model recorded ten behavioral features that indicate the development milestones and are concerned with communication and social behaviors, in addition to the gender, age, ethnicity, some attributes for the family history and the relationship to the person who completed the screening. After training multiple classifiers, three models were shortlisted as candidates for providing exceptionally good accuracy, sensitivity, and specificity. Although the three models' performances are excellent, still the Neural Network is outperforming the other two models, and therefore we propose to implement it. This study showed promising results for ASD screening, and the proposed model can be complemented by developing a mobile application that will add to the convenience and accessibility of the screening method. Furthermore, this research can be improved further and extended by performing feature selection and finding which features are most significant for ASD screening prior to building the models through dimensionality reduction.

## REFERENCES

[1] CDC. Reports, (2020) "AUTISM AND DEVELOPMENTAL DISABILITIES MONITORING (ADDM) NETWORK," ed.

[2] V. Jagan and A. Sathiyaseelan, (2016) "Early intervention and diagnosis of autism," Indian Journal of Health and Wellbeing, vol. 7, no. 12, pp. 1144-1148.

[3] L. E. K. Achenie, A. Scarpa, R. S. Factor, T. Wang, D. L. Robins, and D. S. McCrickard, (2019) "A Machine Learning Strategy for Autism Screening in Toddlers," Journal of Developmental and Behavioral Pediatrics, vol. 40, no. 5, p. 369, doi: 10.1097/DBP.0000000000000668.

[4] P. S. Carbone et al., (2020) "Primary Care Autism Screening and Later Autism Diagnosis," Pediatrics, vol. 146, no. 2, doi: 10.1542/peds.2019-2314.

[5] K. S. Omar, P. Mondal, N. S. Khan, M. R. K. Rizvi and M. N. Islam, (2019) "A Machine Learning Approach to Predict Autism Spectrum Disorder," in 2019 International Conference on Electrical, Computer and Communication Engineering (ECCE): IEEE, pp. 1-6.

[6] S. Broder Fingert et al., (2019) "Implementing systems-based innovations to improve access to early screening, diagnosis, and treatment services for children with autism spectrum disorder: An Autism Spectrum Disorder Pediatric, Early Detection, Engagement, and Services network study," Autism : the international journal of research and practice, vol. 23, no. 3, pp. 653-664, doi: 10.1177/1362361318766238.

[7] C. Kamuk, C. Cantio, and N. Bilenberg, (2017) "Early screening for autism spectrum disorder," European Psychiatry, vol. 41, no. Supplement, pp. S131-S132, doi: 10.1016/j.eurpsy.2017.01.1948.

[8]   L. French and E. M. M. Kennedy, (2018) "Annual Research Review: Early intervention for infants and young children with, or at-risk of, autism spectrum disorder: a systematic review," Journal of Child Psychology and Psychiatry, vol. 59, no. 4, pp. 444-456, doi: 10.1111/jcpp.12828.

[9]   B. Wingfield et al., (2020) "A predictive model for paediatric autism screening," Health informatics journal, p. 1460458219887823, doi: 10.1177/1460458219887823.

[10]  F. Thabtah and D. Peebles, (2020) "A new machine learning model based on induction of rules for autism detection," Health informatics journal, vol. 26, no. 1, pp. 264-286, doi: 10.1177/1460458218824711.

[11]  M. Al-diabat, (2018) "Fuzzy data mining for autism classification of children," International Journal of Advanced Computer Science and Applications, vol. 9, no. 7, pp. 11-17, doi: 10.14569/IJACSA.2018.090702.

[12]  I. M. Nasser, M. O. Al-Shawwa, and S. S. Abu-Naser, (2019) "Artificial Neural Network for Diagnose Autism Spectrum Disorder     " International Journal of Academic Information Systems Research (IJAISR), vol. 3, no. 2, pp. 27-32.

[13]  T. Gennaro, C. Giovanni, D. P. Davide, and A. Stefania, (2020) "Use of Machine Learning to Investigate The Quantitative Checklist For Autism in Toddlers (QCHAT) Towards Early Autism Screening," ed.

[14]  F. Thabtah, (2020) "Autism screening data for     toddlers." https://www.kaggle.com/fabdelja/autism-screening-for-toddlers (accessed.

[15]  R. Loomes, L. Hull, and W. Polmear, (2017) "What is the Male-to-Female Ratio in Autism Spectrum Disorder? A Systematic  Review and Meta-Analysis   " Journal of the American Academy of Child & Adolescent  Psychiatry.

# APPENDIX

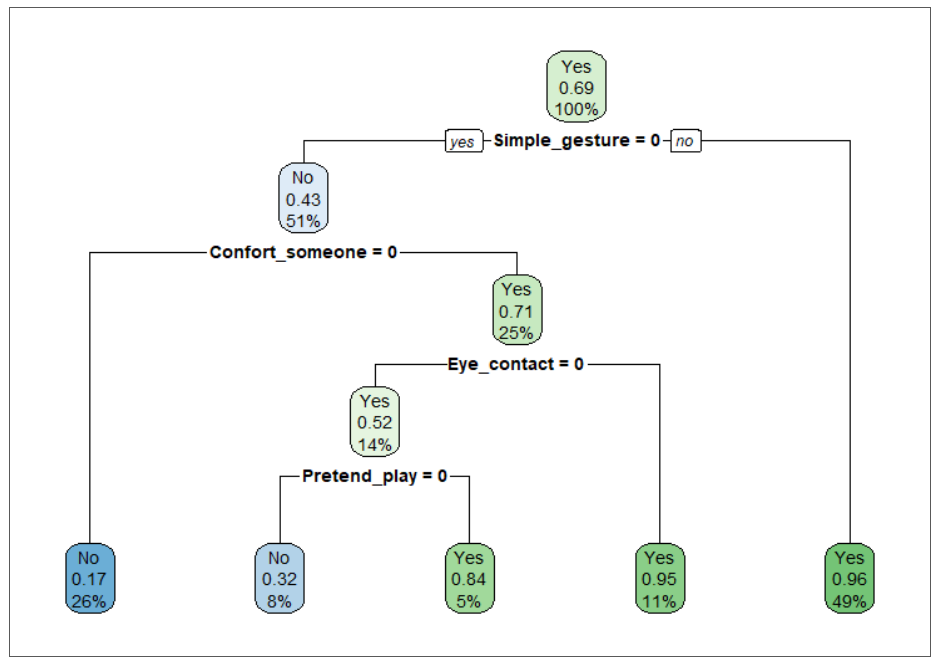

CART Tree

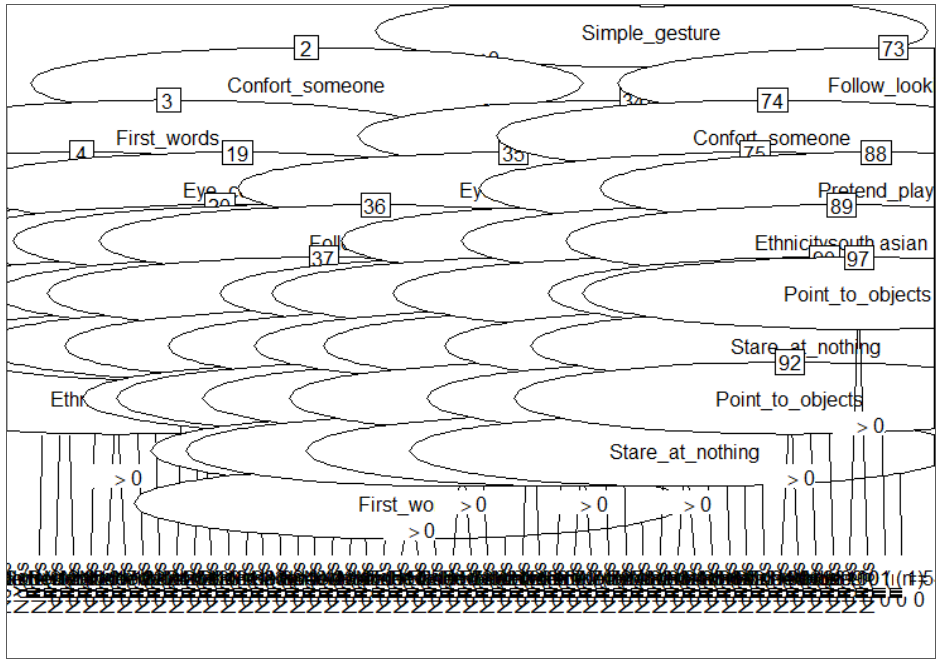

C4.5 Tree

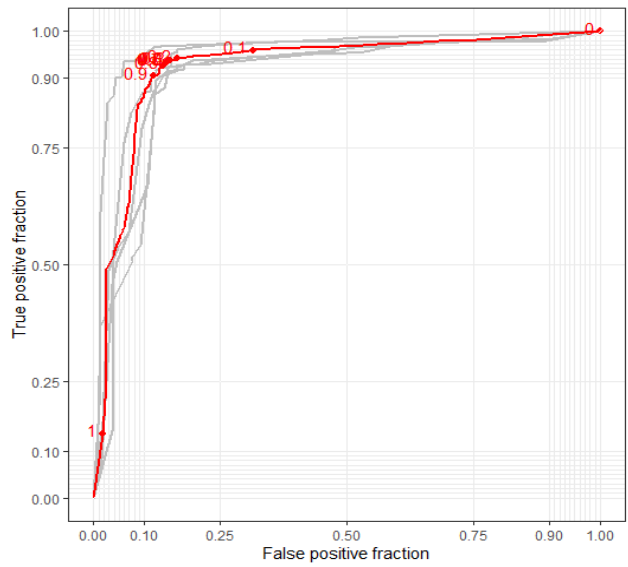

C4.5 ROC curve

Random Forest ROC curve

Neural Network

AUTHORS

**Ms Afef Saihi** is currently working as graduate research and teaching assistant and pursuing her Ph.D. in Engineering Systems management at the American University of Sharjah. Her research interests are in the fields of supply chain management, maintenance planning and optimization, digital transformation and innovation management.

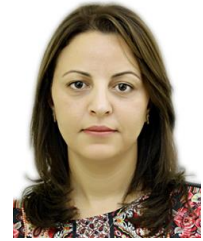

**Dr Hussam Alshraideh** is an Associate Professor of Operations Research and Statistics at the Industrial Engineering Department at the American University of Sharjah (AUS). He holds a dual Ph.D. degree in Industrial Engineering and Operations Research with a minor in Statistics from The Pennsylvania State University. His current research interests include statistical process optimization and smart data analytics applications in healthcare related fields. He has published more than forty papers in highly reputable journals on health informatics and process control.

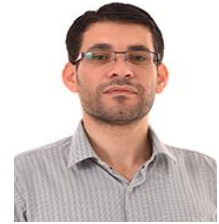